\pdfoutput=1

\documentclass[11pt]{article}

\usepackage{ACL2023}

\usepackage{times}
\usepackage{latexsym}

\usepackage[T1]{fontenc}

\usepackage[utf8]{inputenc}

\usepackage{microtype}

\usepackage{inconsolata}

\usepackage{graphicx}
\usepackage{tabularx}
\usepackage{booktabs}
\usepackage{amssymb}
\usepackage{multirow}
\usepackage{csquotes}
\title{Zero-shot Temporal Relation Extraction with ChatGPT}

\author{Chenhan Yuan, Qianqian Xie, Sophia Ananiadou \\
Department of Computer Science, The University of Manchester \\ \{chenhan.yuan, qianqian.xie, sophia.ananiadou\}@manchester.ac.uk}

\begin{document}
\maketitle
\begin{abstract}
The goal of temporal relation extraction is to infer the temporal relation between two events in the document. Supervised models are dominant in this task. In this work, we investigate ChatGPT's ability on zero-shot temporal relation extraction. We designed three different prompt techniques to break down the task and evaluate ChatGPT. Our experiments show that ChatGPT's performance has a large gap with that of supervised methods and can heavily rely on the design of prompts. We further demonstrate that ChatGPT can infer more small relation classes correctly than supervised methods. The current shortcomings of ChatGPT on temporal relation extraction are also discussed in this paper. We found that ChatGPT cannot keep consistency during temporal inference and it fails in actively long-dependency temporal inference.
\end{abstract}

\section{Introduction}
The temporal relation extraction task aims to extract the temporal relation between either two event triggers in the given document~\cite{dligach-etal-2017-neural}. In this way, a timeline of events in the document can be constructed. It is a crucial task for many downstream NLP tasks, such as natural language understanding~\cite{mani2006machine,paul2017temporal}, storyline construction~\cite{do2012joint,minard2015semeval}, and temporal question answering~\cite{jia2018tequila,jia2018tempquestions}, etc.
Conventionally, recent temporal relation extraction (RE) models are fine-tuned based on pre-trained language models (PLMs), such as BERT and RoBERTa~\cite{devlin2019bert,liu2019roberta}. On top of the PLMs, complex neural networks are applied to classify the temporal relations, such as self-attention~\cite{lin2019bert,ning2018improving}, graph convolutional networks (GCNs)~\cite{mathur2021timers}, and policy network~\cite{man2022selecting}. Most well-performed temporal relation extractors are supervised models, that is, they heavily rely on annotated training documents first before extracting temporal relations on the testing set. However, annotating the temporal relations in training documents requires much domain experts' efforts~\cite{naik2019tddiscourse,ning2018multi}, which is a high cost. 

Different from supervised learning methods, zero-shot learning (ZSL)~\cite{xian2017zero} aims to train the model that can be generalized to unseen data without annotated training data and has attracted much attention in recent years.
Most recently, large language models (LLMs)~\cite{brown2020language,bubeck2023sparks} such as ChatGPT\footnote{\url{https://openai.com/blog/chatgpt}} have exhibited remarkable ability in zero-shot learning on various natural language processing (NLP) and medical tasks~\cite{bang2023multitask}, such as information extraction~\cite{wei2023zero}, machine translation~\cite{jiao2023chatgpt}, summarization evaluation~\cite{luo2023chatgpt}, and mental health detection~\cite{yang2023evaluations}. However, the performance of LLMs in detecting temporal relations between events are not explored yet. Therefore, it is an urgent and spontaneous question if the LLM can perform zero-shot temporal relation extraction tasks well given a proper prompt approach, and if LLM can be the new paradigm for temporal RE. 

In this paper, we explore the performance of ChatGPT on the zero-shot temporal relation extraction, and propose three different prompt strategies to interact with ChatGPT. More specifically, we start with the simple zero-shot prompt that directly requires ChatGPT to infer the temporal relation given the document. Then, we design the event ranking prompt, where ChatGPT is asked to infer the events shown in the given document instead of inferring temporal relations. Finally, we propose the chain-of-thought (CoT) prompt~\cite{weichain} to break down the task into two-stage, which guides ChatGPT to make temporal relation reasoning step by step.  
Based on our experimental results and analysis, we have the following findings:
\begin{itemize}
  \item \textbf{Overall Performance.} ChatGPT significantly underperforms advanced supervised methods and even traditional neural network methods such as Bi-LSTM, indicating the challenge of temporal relation detection with ChatGPT without task-specific fine-tuning.
    \item \textbf{Prompts.} Similar to the finding of recent efforts, the CoT prompt can significantly improve ChatGPT's performance compared with other prompts across all datasets, indicating the importance of proper prompt engineering. 
    \item \textbf{Limitations.} Compared with supervised methods, ChatGPT has better performance in the temporal relations with small proportions in the dataset. However, it is also found to have limitations in detecting long-dependency temporal relation extraction and inconsistent temporal relation inference.
\end{itemize}
\section{Related Work}
\subsection{Temporal Relation Extraction}
Several studies have explored the use of temporal information in relation extraction. For example, Han et al.~\cite{han2019joint,han2019deep} incorporated tense information and timex temporal interactions into their models. Other researchers have proposed using graph neural networks (GNNs) to encode dependency structures, which are important for temporal relation extraction~\cite{mathur2021timers,schlichtkrull2018modeling}. Wang et al.~\cite{wang2022dct} added an attention layer to an R-GCN-based model to focus on document-creation-time (DCT). Man et al.~\cite{man2022selecting} used a reinforcement learning framework to select optimized sentences for input into neural models, improving performance.

In the clinical domain, Leeuwenberg et al.~\cite{leeuwenberg2017structured} applied integer linear programming constraints to learn structured temporal relations. Dligach et al.~\cite{dligach2017neural} improved performance by using neural networks such as CNN and LSTM as the backbone model. Lin et al.~\cite{lin2019bert} utilized pre-trained language models like BERT to learn contextualized embeddings. However, no work has yet explored the feasibility of using LLMs in temporal relation extraction.

\subsection{Zero-shot Learning with ChatGPT}
Since it was launched, ChatGPT has drawn much attention to its strong ability for various NLP tasks. 
In the clinical domain, Tang et al. explored ChatGPT's ability on zero-shot named entity recognition and relation extraction~\cite{tang2023does}. The experiments on NCBI Disease and BC5CDR Chemical datasets showed that ChatGPT cannot recognize named entities correctly in the clinical domain as the F1 drops around 55.94\%-91.58\% compared to SOTA-supervised methods. ChatGPT's poor clinical NER was also proved by testing in i2b2 dataset~\cite{hu2023zero}. However, it can achieve comparable performance in clinical relation extraction as the F1 score only decreased by 4.73\%-10.93\%~\cite{tang2023does, agrawal2022large}.

In extraction-related tasks, some work evaluated ChatGPT's ability in event extraction, general information extraction, and relation extraction~\cite{tang2023does,gao2023exploring,wei2023zero}. The evaluation process follows multi-stage interactions/conversations with ChatGPT and guides it to produce the desired answers. The evaluation results of these studies showed that with proper prompting, ChatGPT can achieve comparable performance with the supervised methods on zero-shot or few-shot settings of extraction tasks. However, there is also some work pointing out that ChatGPT's ability is still limited in some specific extraction scenarios such as extracting clinical notes with privacy information masked~\cite{tang2023does}. Some work also discussed that for non-digital-available texts, such as historical documents, the entity recognition was performed poorly by ChatGPT~\cite{gonzalez2023yes,borji2023categorical}. 

\section{ChatGPT for Zero-shot Temporal Relation Extraction}
\begin{figure*}[t]
\centering
\includegraphics[width=0.95\textwidth]{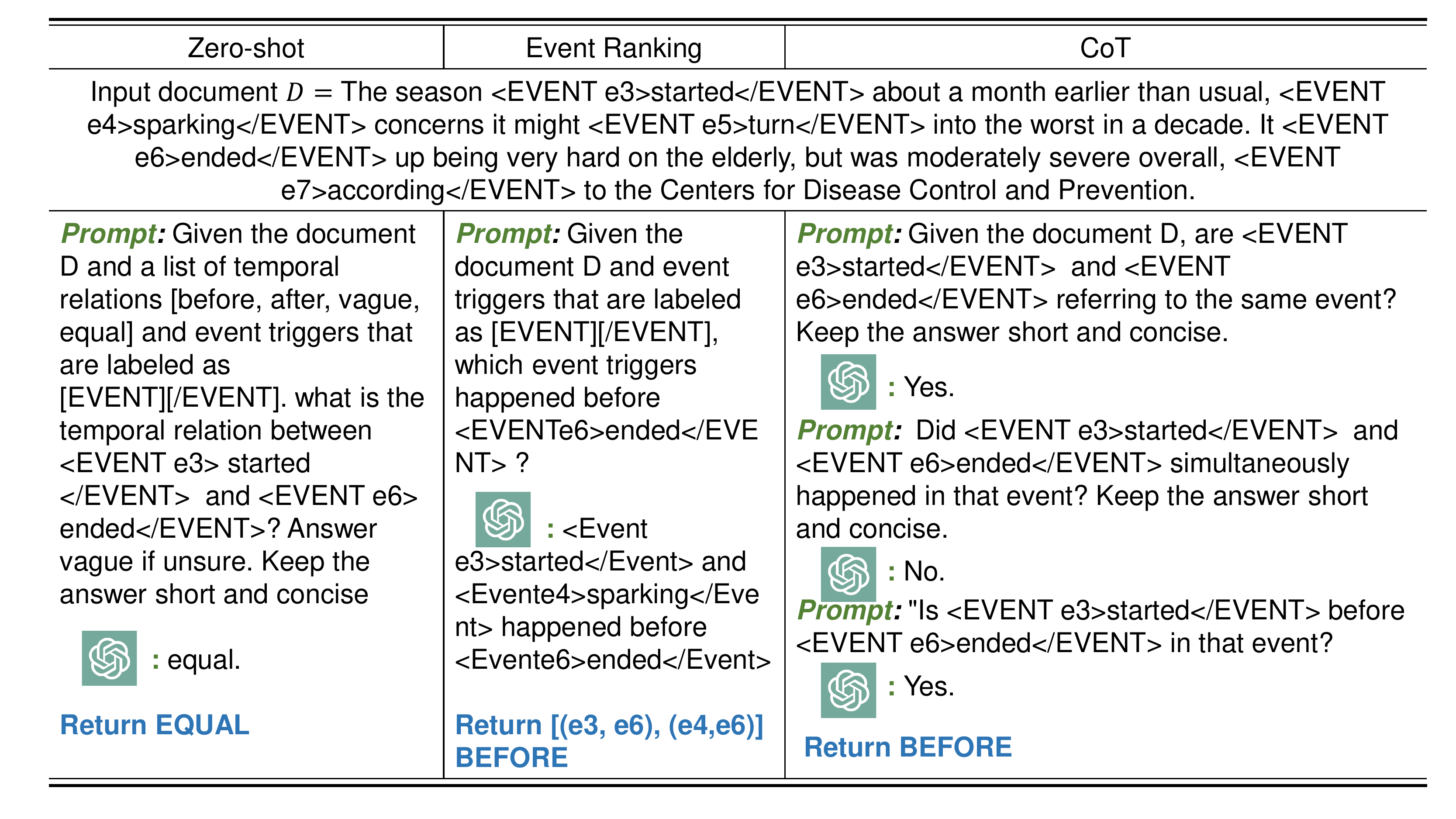}
\caption{The proposed prompts.}
\label{fig:chat_exp}
\end{figure*}
Given the input document, the temporal relation extraction aims to identify the temporal relation between any two event triggers in the document, which is modeled as the multi-classification problem. We propose three different prompts methods to evaluate ChatGPT's performance on temporal relation extraction as shown in Figure~\ref{fig:chat_exp}. 
\subsection{Zero-shot Prompt}
In this prompt, given the document $D=\{x_1,x_2,x_3,\cdots,x_n\}$, we first label the event triggers with <EVENT></EVENT>. That is if $x_i$ is an event trigger, we label it as <EVENT>$x_i$</EVENT>. Then the whole labeled document is sent to ChatGPT and we ask ChatGPT to find the temporal relations between either two events. 
Note that our goal is to test ChatGPT's zero-shot ability in the temporal relation extraction task. Therefore, we do not provide any examples in our prompts. As shown in Figure.~\ref{fig:chat_exp}, we give ChatGPT the whole document, the list of all temporal relations, and the annotations of event triggers. In the end, the input question is designed as ``what is the temporal relation between <EVENT>$x_i$</EVENT> and <EVENT>$x_j$</EVENT>?''. We let ChatGPT to answer the question by using the temporal relations provided in the list.

\subsection{Event Ranking Prompt}
\begin{figure}[h]
\centering
\includegraphics[width=0.45\textwidth]{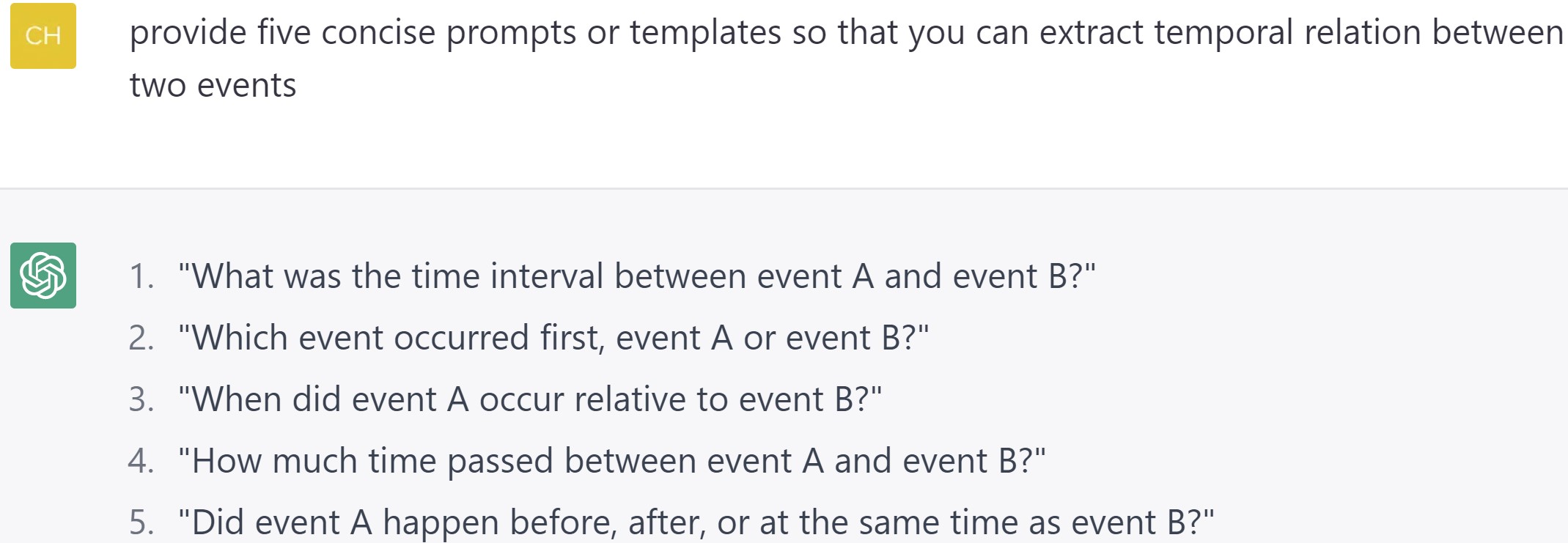}
\caption{Prompts generated by ChatGPT.}
\label{fig:chat_prompt}
\end{figure}

Further, we design a new prompt to make the task easier to learn for ChatGPT. Specifically, given the document $D$ and one event trigger $e_i$ with the form <EVENT>$x_i$</EVENT> and temporal relation set $R$, we require ChatGPT to complete the query $(e_i, r_j, ?) \: \forall r_j\in R$. In this way, instead of querying  $(e_i, ?, e_j)$ as in the previous prompt, ChatGPT is required to predict the missing event trigger. As event triggers are already shown in the given text/document, ChatGPT is more likely to infer the event triggers than unseen temporal relations as they are not explicitly provided in the context.
In detail, as shown in Fig.~\ref{fig:chat_exp}, we achieve this by asking the question first, such as ``Which events happened before $e_i$?''. Then based on the feedback of ChatGPT, we form the $<e_i,r,e_j>$ triplets to perform the evaluation. Note that if the same event pair is detected in different temporal relation (>2), we denote that this event pair has a ``vague'' relation as ChatGPT cannot confidently determine which temporal relation the event pair is classified. We also asked ChatGPT to provide some concise prompts to perform temporal relation extraction tasks. As shown in Fig~\ref{fig:chat_prompt},  the prompts provided by ChatGPT are in line with our event ranking prompt approach.

\subsection{Chain-of-thought Prompt}
We notice that if two event triggers refer to the same event but point to different timestamps of the duration of that event, ChatGPT cannot distinguish them. For example, in Figure.~\ref{fig:chat_exp}, ``<EVENT e3> started'' and ``<EVENT e6> ended'' referred to the beginning and the end of the season event. ChatGPT assumes that these two event triggers happened at the same time if we directly ask about the temporal relation following the zero-shot prompt. Therefore, we propose a new chain-of-thought prompt with two steps, which firstly navigates ChatGPT to distinguish event triggers referring to the same event, and then guides ChatGPT to infer their temporal relation.
Specifically, given the document $D$ and two event triggers $e_1$ and $e_2$, we first ask ChatGPT to determine if $e_1$ and $e_2$ refer to the same event. If they are not, we further ask ChatGPT to determine the temporal relation between the two event triggers. If they point to the same event, we use a similar prompt but with the extra phrase ``in that event'' to ask ChatGPT. As shown in Fig.~\ref{fig:chat_exp}, we first ask ChatGPT if the two event triggers <EVENT>started</EVENT> and <EVENT>ended</EVENT> refer to the same event. Then based on ChatGPT's feedback, we further iteratively go through the whole temporal relation list to determine which temporal relation exists between the two event triggers. We show the full prompts we designed in detail in Appendix~\ref{appx:pro}.
\section{Experiments}
\subsection{Datasets}
We use three datasets to evaluate the performance of ChatGPT on the zero-shot temporal relation extraction. The statistical details of these datasets are shown in the Table~\ref{table:stats}.
\begin{table}[htbp]
\centering
\begin{tabular}{lcccc}
\toprule
Dataset        & Train  & Dev   & Test  & Labels \\ \midrule
TB-Dense & 4,032  & 629   & 1,427 & 6      \\
MATRES         & 6,336    & --    & 837    & 4      \\
TDDMan         & 4,000  & 650   & 1,500 & 5      \\ \bottomrule
\end{tabular}
\caption{Statistics of the number of annotated event pairs and different temporal relation labels of the MATRES, TB-Dense and TDDMan datasets. }
\label{table:stats}
\end{table}

\begin{itemize}
    \item \textbf{TimeBank-Dense}~\cite{cassidy2014annotation}  labeled 36 news documents in total. The temporal relations between event triggers-event triggers, timex-event triggers, timex-timex, are labeled. Following previous work, we only test our model on the event trigger-event trigger relation in the testing set. 
    \item \textbf{MATRES}~\cite{ning2018multi} is a dataset primarily focusing on temporal relations of event triggers with local sentences (1 or 2 sentences). 
    \item \textbf{TDDiscourse}~\cite{naik2019tddiscourse} was created to explicitly emphasize global discourse-level temporal ordering. Based on the annotation accuracy, the dataset is split into TDDMan and TDDAuto, where TDDAuto introduces much more automatic labels and noise. In this paper, we only evaluate ChatGPT on TDDMan because of the budget limitation.
\end{itemize}
We only use the testing set of each dataset directly to test our approaches as we do not require training of ChatGPT, following the zero-shot setting. We report the F1 score on each dataset and each temporal relation. 
\subsection{Baseline Models}
Since there is no zero-shot learning methods for temporal relation extraction before, we compare the performance of ChatGPT with the following advanced supervised methods:
\begin{itemize}
    \item CAEVO~\cite{chambers2014dense} a sieve-based architecture that includes multi-level classifiers for intra-sentence temporal relation learning.
    \item SP+ILP~\cite{ning2017structured} a structured learning approach that captures the global temporal features when inferring the relation between two local events.
    \item Bi-LSTM~\cite{cheng2017classifying} a Bi-LSTM-based model that encodes the dependency path between two events to classify temporal relation.
    \item Joint~\cite{han2019joint} a joint end-to-end event and temporal relation extraction model that shares the contextualized embedding of two sub-tasks
    \item Deep~\cite{han2019deep} a neural network model that utilized SSVM as the scoring function to learn temporal constraints and context embeddings.
    \item UCGraph~\cite{ijcai2021p533} a graph-based model that is trained with mask pre-training mechanism. The model's uncertainty level is used to guide the inference during testing.
    \item TIMERS~\cite{mathur2021timers} a graph-based model that leverages three graphs to learn temporal, rhetorical, and syntactic information, respectively.
    \item SCS-EERE~\cite{man2022selecting} a reinforcement learning-based selector is designed to select the optimized sentences for temporal inference between the given two events.
    \item RSGT~\cite{zhou2022rsgt}  a syntactic-and-semantic-based graph model pre-trained on a temporal neighbor prediction task.
    \item FaithTRE~\cite{wang2022extracting} a model that applied Dirichlet prior to estimating the correctness likelihood. A temperature scaling is also used to recalibrate the model confidence measure after bias mitigation.
    \item DTRE~\cite{wang2022dct} a document creation time(DCT)-aware graph with a global consistency mechanism when inferring temporal relations.
    \item MulCo~\cite{yao2022multi} a joint model using the BERT to learn contextualized features and GNN to capture syntactic structures. The two models are combined via a multi-level contrastive learning framework.
\end{itemize}

\subsection{Results}
\begin{table*}[t]
\centering
\begin{tabular}{lccccccccc}
\toprule
\multirow{2}{*}{Models} & \multicolumn{3}{c}{MATRES} & \multicolumn{3}{c}{TDDMan} & \multicolumn{3}{c}{TB-Dense} \\ \cmidrule{2-10} 
                          & prec     & recall     & F1      & prec      & recall      & F1       & prec      & recall      & F1     \\ \midrule
CAEVO~\cite{chambers2014dense} &--&--&--&32.3 &10.7 &16.1 & 49.9 &46.6 &48.2\\
SP+ILP~\cite{ning2017structured}  &71.3 &82.1 &76.3& 23.9 &23.8 &23.8 & 58.4 &58.4 &58.4\\
Bi-LSTM~\cite{cheng2017classifying} &59.5 &59.5 &59.5 &24.9 &23.8 &24.3 & 63.9 &38.9 &48.4\\
Joint~\cite{han2019joint} & -- & -- &75.5& 41.0 & 41.1 & 41.1 & -- &-- & 64.5\\
Deep~\cite{han2019deep} &77.4 &86.4 &81.7 &--&--&--&62.7 &58.9 &62.5\\
UCGraph~\cite{ijcai2021p533}  & -- & -- & -- & 44.5 & 42.3 & 43.4 & 62.4 & 56.1 & 59.1\\
TIMERS~\cite{mathur2021timers}  & 81.1 & 84.6 & 82.3 & 43.7 & 46.7 & 45.5 & 48.1 & 65.2 & 67.8\\  
SCS-EERE~\cite{man2022selecting}       &  78.8 & 88.5 & 83.4 &  -- & -- & 51.1 &  -- & -- & -- \\
FaithTRE~\cite{wang2022extracting}   &--&--& 82.7 &--&--&52.9 & -- & -- &--\\ 
RSGT~\cite{zhou2022rsgt}       &  82.2 & 85.8 & 84.0 &  -- & -- & -- &  68.7 &68.7 &68.7 \\
DTRE~\cite{wang2022dct}   &--&--&--& 56.3 &56.3 &56.3 & --&--&70.2\\
MulCo~\cite{yao2022multi}  &88.2 &88.2 &88.2 &56.2 &54.0 &55.1  &84.9 &84.9 &84.9\\
\midrule
\midrule
ChatGPT\_{ZS}      & 26.4     & 24.3       & 25.3  & 17.7    & 13.6 & 15.3   &  23.7 &14.3 &  17.8   \\
ChatGPT\_{ER}   &   21.9  & 17.3 & 19.3 &   3.7 &   0.3 &   0.5  & 37.6 & 35.8 &  36.6    \\
ChatGPT\_{CoT} & 48.0  & 57.7 &  52.4   &26.8 &22.3 & 24.3  &  43.4 &32.2&  37.0   \\ \bottomrule
\end{tabular}
\caption{The comparison of ChatpGPT with various prompt techniques and supervised state-of-the-art models.}
\label{table:model_comp}
\end{table*}
In table~\ref{table:model_comp}, we can see that ChatGPT struggles to outperform supervised state-of-the-art models such as SCS-EERE~\cite{man2022selecting} and RSGT~\cite{zhou2022rsgt}, and even the traditional neural networks methods such as CAEVO and BI-LSTM, indicating its ineffective in the temporal relation extraction task in the zero-shot setting. Table~\ref{table:model_comp} also shows the performance of ChatGPT under different prompts in three datasets. We noticed that ChatGPT\_{ER} yields the worst performance on the MATRES and TDDMan datasets. ChatGPT's performance with the event ranking prompt on TDD-Man dataset is poor as most event trigger pairs cannot be detected under this prompt. However, if the event trigger pairs are explicitly fed to ChatGPT, it can somehow partially infer the temporal relations correctly. For example, the zero-shot prompt and CoT prompt could improve the F1 score by $14.8\%$ and $23.8\%$, respectively.
While for the TB-Dense dataset, ChatGPT\_{ZS} has the worst performance. ChatGPT\_{CoT} achieves the best performance across all datasets, such as it significantly outperforms ChatGPT\_{ZS} and ChatGPT\_{ER} by 27.1\% and 33.1\%.
This illustrates the effectiveness of the CoT prompt with step-by-step guidance in prompting ChatGPT.
\begin{table*}[t]
\centering
\begin{tabular}{lccccccccc||ccc}
\toprule
\multirow{2}{*}{Relation} & \multicolumn{3}{c}{Zero-shot} & \multicolumn{3}{c}{CoT} & \multicolumn{3}{c}{Event ranking} & \multicolumn{3}{c}{Deep} \\ \cmidrule{2-13} 
                          & prec     & recall     & F1      & prec      & recall      & F1       & prec      & recall      & F1     & prec      & recall      & F1\\ \midrule
overall                   & 26.4     & 24.3       & 25.3    & 48.0      & 57.7        & 52.4     &   21.9       &   17.3   &  19.3 &77.4& 86.4 & 81.7 \\
EQUAL       & 0.0      & 0.0        & 0.0     & 7.1       & 2.9         & 4.1      &   5.8  &   11.1  &  7.6   & 0.0 & 0.0 & 0.0   \\
VAGUE     & 14.3     & 58.7       & 23.1    & 14.4      & 8.1         & 10.4     &    14.6  &  86.2 & 25.0  & 0.0 & 0.0 & 0.0  \\
AFTER   & 34.0     & 25.6       & 29.2    & 41.6      & 41.8        & 41.7     &   36.4        &    1.6    & 3.0  & 72.3 & 84.8 & 78.0     \\
BEFORE  & 52.5     & 17.8       & 26.6    & 63.1      & 71.6        & 67.1     &   57.0  &  13.0 &  21.1  & 80.1 & 89.6 & 84.6\\ \bottomrule
\end{tabular}
\caption{The zero-shot performance of ChatGPT with three different prompts on the MATRES dataset.}
\label{table:matres}
\end{table*}

\begin{table*}[t]
\centering
\begin{tabular}{lccccccccc}
\toprule
\multirow{2}{*}{Relation} & \multicolumn{3}{c}{Zero-shot} & \multicolumn{3}{c}{CoT} & \multicolumn{3}{c}{Event ranking} \\ \cmidrule{2-10} 
                          & prec     & recall     & F1      & prec      & recall      & F1       & prec      & recall      & F1     \\ \midrule
overall                   & 17.7    & 13.6       &15.3     & 26.8     & 22.3       & 24.3     & 3.7     & 0.3       & 0.5  \\
is included       & 9.5     & 0.7        & 1.3     & 20.9      & 3.1        & 5.4       & 0.0      & 0.0        & 0.0       \\
include     & 41.9     & 17.7     & 24.8    & 37.9      & 11.2        & 17.3      & 0.0      & 0.0        & 0.0   \\
after   & 14.7    & 9.0      & 11.2  & 33.3      & 4.3        & 7.5      & 0.0      & 0.0        & 0.0      \\
before  & 29.7    & 22.9       & 25.9   & 35.1    &70.8       & 46.9     & 12.5     & 0.7       & 1.4 \\
simultaneous  & 3.9    & 39.1       & 7.0    & 0.0     & 0.0       & 0.0     & 11.1     & 2.2       & 3.6
\\ \bottomrule
\end{tabular}
\caption{The zero-shot performance of ChatGPT with three prompts on the TDD-Man dataset.}
\label{table:tddman}
\end{table*}

Table~\ref{table:matres}, Table~\ref{table:tddman} and Table~\ref{table:tbd} further list the detail results of ChatGPT with different prompts on three datasets. We can see that the performance with the event ranking prompt is much better than that on other datasets. ChatGPT with the zero-shot prompt cannot determine the temporal relation ``is included'' and therefore yields a 0.0 performance in this type of relation. The CoT prompt improves the overall performance by significantly detecting ``before'' and ``after'' temporal relations. As these two relations take a great portion of the whole dataset, the overall performance is also improved.
\begin{table*}[t]
\centering
\begin{tabular}{lccccccccc}
\toprule
\multirow{2}{*}{Relation} & \multicolumn{3}{c}{Zero-shot} & \multicolumn{3}{c}{CoT} & \multicolumn{3}{c}{Event ranking} \\ \cmidrule{2-10} 
                          & prec     & recall     & F1      & prec      & recall      & F1       & prec      & recall      & F1     \\ \midrule
overall                   & 23.7    & 14.3       &17.8     & 43.4     & 32.2      & 37.0     & 37.6    & 35.8      & 36.6 \\
is included       & 0.0     & 0.0        & 0.0     & 10.0      & 1.9        & 3.2       & 6.2      & 3.8        & 4.7       \\
include     & 3.3     & 10.7     & 24.8    & 5.5      & 16.1        & 8.2      & 16.7    & 5.4       & 8.1   \\
after   & 29.0    & 17.2      & 11.2  & 70.4      & 13.9       & 23.2      & 19.0     & 1.5      & 2.7    \\
before  & 40.0    & 9.9       & 25.9   & 35.0    &75.5       & 47.9     & 31.2     & 25.3      & 27.9 \\
simultaneous  & 1.5    & 45.5       & 3.0    & 33.3     & 4.5       & 8.0     & 6.7     & 50.0    & 11.8\\
vague  & 44.6    & 24.0       & 31.2    & 51.2     & 29.6      & 37.5     & 46.0     & 63.5       & 53.4
\\ \bottomrule
\end{tabular}
\caption{The zero-shot performance of ChatGPT with three prompts on the TimeBank-Dense dataset.}
\label{table:tbd}
\end{table*}

\section{Discussion}
\subsection{ChatGPT is slightly better on small temporal relation classes}
The imbalanced data is a severe long-existing problem in the temporal relation extraction task. Because of the events' temporal order frequency in real life, some temporal relations, such as  ``simultaneous'' and ``equal'', are very limited in most temporal relation extraction datasets. And popular NLP data-augmented methods are difficult to be applied in the temporal domain. Therefore, most state-of-the-art supervised methods yield much worse performance on small relation classes.

In Deep~\cite{han2019deep}, the authors reported detailed performance on each temporal class in MATRES dataset. We therefore especially compare the performance on small relation classes against Deep, i.e., ``EQUAL'' and ``VAGUE''. As shown in Table~\ref{table:matres}, the supervised model Deep could achieve much better overall performance ($81.7$ F1 score), due to the contribution of two majority relations, ``before'' and ``after''. Compared to Deep's 0.0 performance on ``EQUAL'' and ``VAGUE'', ChatGPT with event ranking, CoT and zero-shot prompts can correctly extract some small class relations.

\subsection{ChatGPT failed in actively long-dependency temporal relation extraction}
As shown in Table~\ref{table:model_comp}, ChatGPT's performance drops a lot in the TDDMan dataset. We argue that this is mainly because the TDDMan focuses more on discourse-level temporal relations and ChatGPT failed to extract useful information from long documents. 
As shown in Table~\ref{table:tddman}, the event ranking prompt yields almost 0.0 on the whole dataset. In practice, we initially input the whole document $D$ to ChatGPT and ask ChatGPT which event triggers in the document $D$ happened before the given event trigger $e_1$. That is, ChatGPT should actively search all event triggers in the document and produce answers. However, in the TDDMan dataset, ChatGPT cannot produce a formatted answer and the outputs sometimes are even some random words in the document instead of event triggers. We then test the limit of the size of the input document, i.e., the number of sentences. We finally found that if we limit the size of the input document to at most 8 context sentences around the event triggers, ChatGPT would be more stable to produce formatted answers instead of repeating part of the document randomly. However, in this way, most temporal relations extracted by ChatGPT do not match with the golden labels in the TDDMan dataset because the extracted temporal relations are only in short dependency while TDDMan emphasizes long-distance temporal dependency.

Nevertheless, surprisingly, if we explicitly ask ChatGPT what the temporal relations between two event triggers labeled in the document are, ChatGPT can answer some of them correctly. Note that we do not cut the size of the document in this case and ChatGPT can still learn some of the temporal dependency in the document. Compared to the event ranking prompt, ChatGPT passively receives two event trigger information with the zero-shot and CoT prompts. This may reduce the inference difficulty as more information is given.

\subsection{ChatGPT can be improved via multi-stage ``yes'' or ``no'' prompts}
Intuitively, the most efficient way to query ChatGPT about temporal relations between two events in the given document should be the zero-shot prompt, which directly asks the ChatGPT to answer the temporal relation between any two event triggers. 
If only one event trigger is given, then the prompts provided by ChatGPT, i.e., event ranking, should be used to interact with ChatGPT. 
However, our extensive experiments show that these two prompt methods produce much worse performance in most cases compared to the CoT prompt. Note that both zero-shot and CoT provide sufficient information about event triggers. We argue there is another difference resulting in the performance gap.

Comparing the two prompts, one significant difference is that the CoT prompt only accepts ``yes'' or ``no'' answers while the zero-shot prompt returns a specific temporal relation label. In the zero-shot prompt, ChatGPT is required to select a temporal relation from the given list, which is similar to a conventional multi-class classification problem. However, in the CoT prompt, ChatGPT only has to determine if one specific temporal relation exists (or not) between two event triggers. This simplified the problem into a binary classification. Further, with the previous question-answer pair as context, ChatGPT has a higher probability of making the correct selection. For example, in Fig~\ref{fig:chat_exp}, in the first round, ChatGPT already inferred that the relation is not ``EQUAL''. Then in the second round, ChatGPT is more confident to predict the temporal relation from \emph{[BEFORE, AFTER, VAGUE]} instead of ``EQUAL''.

\subsection{ChatGPT's temporal inference is inconsistent even with sufficient context}
\begin{figure*}[t]
\centering
\includegraphics[width=0.95\textwidth]{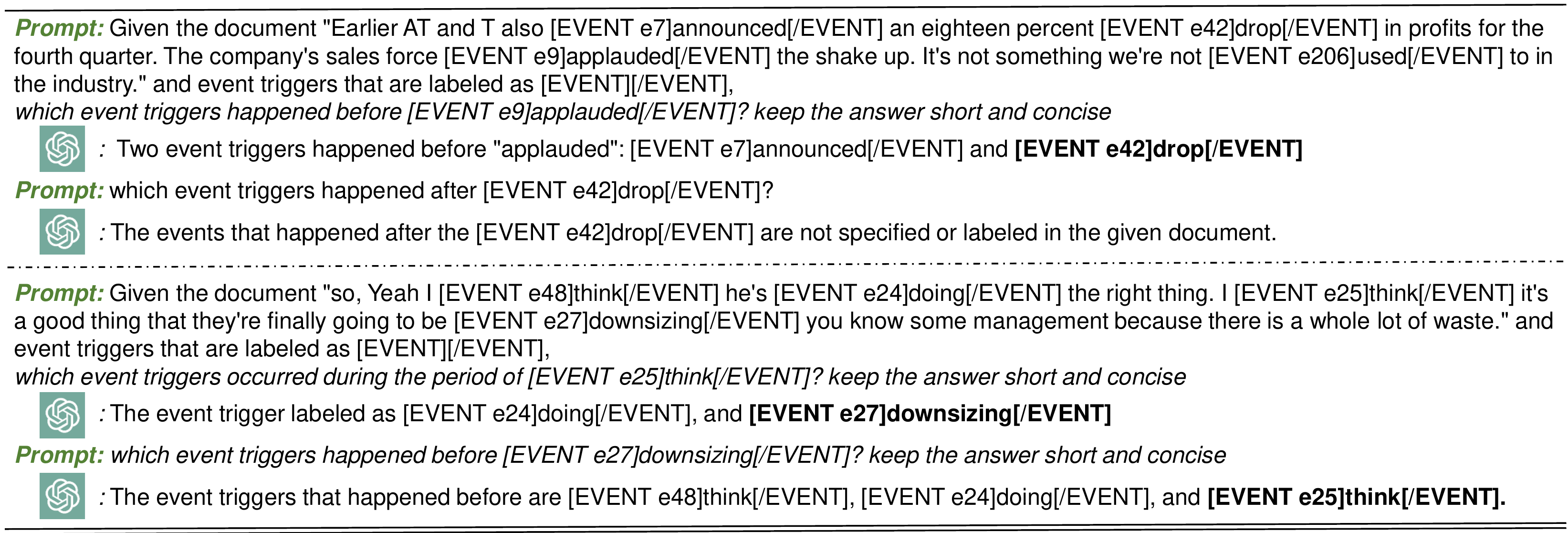}
\caption{ChatGPT's two temporal inconsistency cases examples in the Event ranking prompt.}
\label{fig:chat_event_rank}
\end{figure*}
During the testing of ChatGPT with the event ranking prompt, we noticed a fatal issue in ChatGPT's temporal relation extraction, namely the inconsistent temporal relation inference. Given the same input document, ChatGPT may produce different temporal relations between two event triggers. As the top example shown in Figure~\ref{fig:chat_event_rank}, given the document $D$, if the prompt is ``Which event triggers happened \emph{before} $e_1$?'', ChatGPT will give a list of event triggers, e.g, [$e_2,e_5,e_6$]. Now given the same document $D$, if the prompt is ``Which event triggers happened \emph{after} $e_6$'', ChatGPT is expected to at least include $e_1$ in the list. However, during the experiments, we noticed that ChatGPT failed in this scenario multiple times and the failures can be categorized into two cases. The first failure case is that ChatGPT does not include $e_1$ in any list associated with $e_6$. The second case is that ChatGPT includes $e_1$ in a wrong list, e.g., ``EQUAL'', associated with $e_6$ which violates the temporal consistency.

\begin{figure*}[t]
\centering
\includegraphics[width=0.95\textwidth]{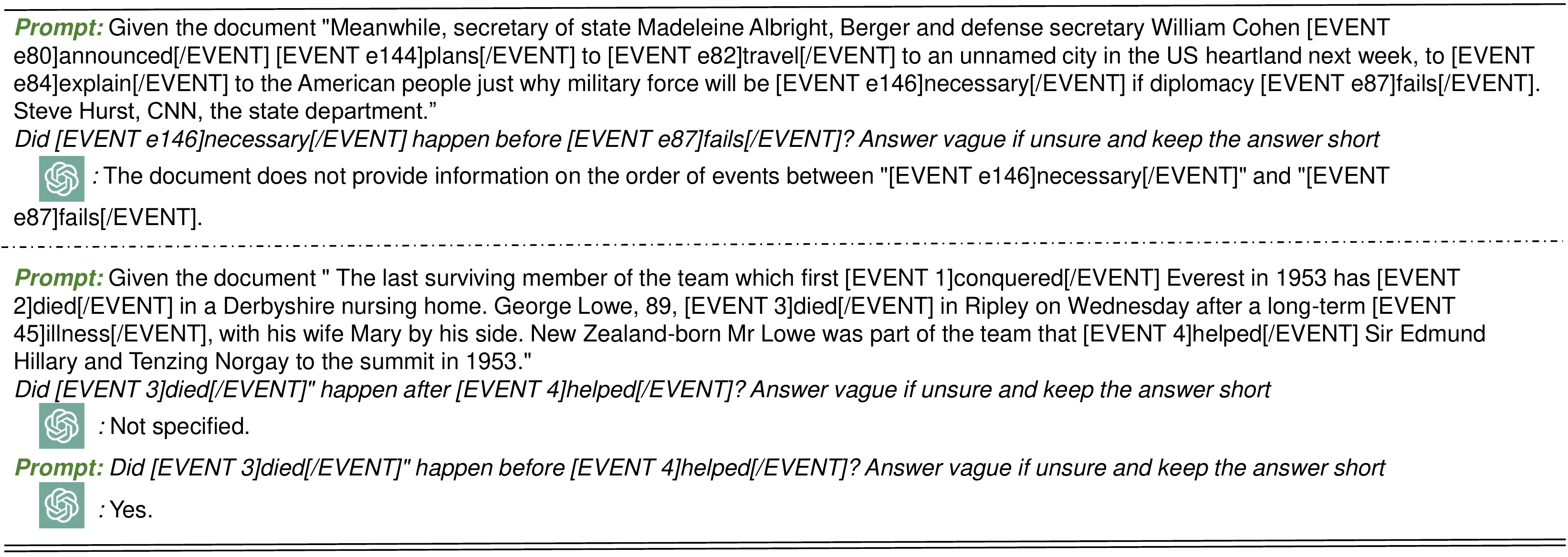}
\caption{ChatGPT's inconsistency failures examples in the CoT prompt. The top shows ChatGPT does not reply ``vague'' when unsure and the bottom shows that ChatGPT still infers other inconsistent relations.}
\label{fig:chat_unknown}
\end{figure*}

A similar problem also happened to the CoT prompt. As shown in Figure~\ref{fig:chat_unknown}, during the experiments, ChatGPT would give various answers instead of ``vague'' as we specified, if it thinks there is no clue to infer the temporal relation between the given two event triggers. These answers include ``Cannot determine.'', ``I cannot answer that question as it is unclear from the given information.'', ``Unknown.'', etc. We treat all of these answers as ``vague'' when we evaluate ChatGPT. Intuitively, if ChatGPT cannot answer ``yes'' or ``no'' for the specific temporal relation between two event triggers, it should also not be able to answer questions about other temporal relations of the same event triggers. However, ChatGPT may violate its ``unknown'' statement in the multi-stage prompts. We test if ChatGPT's inconsistency exists in multi-stage prompts by implementing the following experiments. For example, during the $i$-th round, if we ask ``Did $e_1$ happen before $e_2$?'' and ChatGPT's answer includes ``It is unclear from the given information.''. We then ask the next $i+1$ question ``Did $e_1$ happen after $e_2$?''. And surprisingly, in most cases ($84\%$), ChatGPT can classify the event trigger pair into other temporal relation classes even if it claims that the information is insufficient. Moreover, $96\%$ of these classification results is incorrect. 
\section{Conclusion}
In this work, we comprehensively test ChatGPT's zero-shot ability on temporal relation extraction. We designed three different prompts to evaluate ChatGPT's performance. Our extensive experiments demonstrate that with proper prompting, ChatGPT's performance on zero-shot temporal relation extraction can be significantly improved, highlighting the importance of prompt engineering to better trigger ChatGPT's ability in future work. However, compared to supervised methods, ChatGPT is still far behind. We further discuss our findings from experimental results, including its better performance in small classes than supervised methods and its drawbacks such as failures in long-distance temporal dependency inference and inconsistent temporal relation inference.
Our work only takes the initial step of exploring the LLMs for zero-shot temporal relation extraction.
To fill the gap between the performance of LLMs in the zero-shot setting and that of advanced supervised methods, we believe more efforts should be explored in the future, such as the few-shot prompt engineering and investigating the performance of other LLMs such as GPT-4~\cite{bubeck2023sparks}.
\bibliography{anthology,custom}
\bibliographystyle{acl_natbib}

\appendix
\section{Detailed Prompts}
\label{appx:pro}
In this section, we provide detailed prompts we used in the experiments.

For datasets with the following temporal relations, such as MATRES: \emph{[before, after, vague, equal]}, the following prompt is used for Zero-shot method:
\begin{displayquote}
Given the document $D$ and a list of temporal relations [before, after, vague, equal] and event triggers are labeled as [EVENT][/EVENT]. What is the temporal relation between $e_1$ and $e_2$? Answer vague if unsure. keep the answer short and concise
\end{displayquote}
The prompts designed for Event ranking are shown as follows:
\begin{displayquote}
Given the document $D$ and event triggers that are labeled as [EVENT][/EVENT], which event triggers happened ``relation'' $e_1$? keep the answer short and concise \{If the relation that will be asked is ``before'' or ``after''\}

Given the document $D$ and event triggers that are labeled as [EVENT][/EVENT], which event triggers were happening at the same time as $e_1$? keep the answer short and concise \{If the relation that will be asked is ``equal''\}
\end{displayquote}
We propose the following prompts for CoT method:
\begin{displayquote}
 Is $e_1$ ``relation'' $e_2$? Keep the answer short and concise. \{If the relation that will be asked is ``before'' or ``after''\}
 
Did $e_1$ and $e_2$ simultaneously happen? Keep the answer short and concise. \{If the relation that will be asked is ``equal''\}

Is the temporal relation of $e_1$ and $e_2$ vague? Keep the answer short and concise. \{If the relation that will be asked is ``vague''\}
\end{displayquote}

Some datasets accompany by the following temporal relations, such as TB-Dense and TDDiscouse: \emph{[before, after, vague, include, is included, simultaneous]}. Note that ``vague'' can be optional for some datasets, but we still ask ChatGPT to produce a ``vague'' label if it cannot infer temporal relation in the given two events. The following prompt is used for the Zero-shot method:
\begin{displayquote}
Given the document $D$ and a list of temporal relations [before, after, include, is included, simultaneous] and event triggers are labeled as [EVENT][/EVENT]. what is the temporal relation between $e_1$ and $e_2$? Answer vague if unsure. keep the answer short and concise
\end{displayquote}
The prompts designed for Event ranking are shown as follows:
\begin{displayquote}
Given the document $D$ and event triggers that are labeled as [EVENT][/EVENT], which event triggers happened ``relation'' $e_1$? keep the answer short and concise \{If the relation that will be asked is ``before'' or ``after''\}

Given the document $D$ and event triggers that are labeled as [EVENT][/EVENT], which event triggers were happening at the same time as $e_1$? keep the answer short and concise \{If the relation that will be asked is ``simultaneous''\}

Given the document $D$ and event triggers that are labeled as [EVENT][/EVENT], which event triggers were included in $e_1$? keep the answer short and concise \{If the relation that will be asked is ``include''\}

Given the document $D$ and event triggers that are labeled as [EVENT][/EVENT], which event triggers that $e_1$ were included in? keep the answer short and concise \{If the relation that will be asked is ``is included''\}
\end{displayquote}
In the CoT method, we first provide the following prompt: 
\begin{displayquote}
    Given the document $D$, are $e_1$ and $e_2$ referring to the same event? Keep the answer short and concise
\end{displayquote}
Then the following prompts are fed to ChatGPT if the answer tends to ``NO'':
\begin{displayquote}
 Did $e_1$ happen ``relation'' $e_2$? Keep the answer short and concise. \{If the relation that will be asked is ``before'' or ``after''\}
 
Did $e_1$ and $e_2$ simultaneously happen? Say yes or no and keep the answer short and concise"
 \{If the relation that will be asked is ``simultaneous''\}

Did $e_2$ occur during the time period of $e_1$? Keep the answer short and concise \{If the relation that will be asked is ``include''\}

Did $e_1$ occur during the time period of $e_2$? Keep the answer short and concise \{If the relation that will be asked is ``is included''\}
\end{displayquote} Note that if ChatGPT determines that the two event triggers refer to the same event, we add a phrase ``in that event'' in the end of each prompt mentioned above.
\end{document}